\pdfoutput=1

\documentclass[11pt]{article}

\usepackage[]{acl}

\usepackage{times}
\usepackage{latexsym}

\usepackage[T1]{fontenc}

\usepackage[utf8]{inputenc}
\usepackage{graphicx}
\usepackage{microtype}
\usepackage{tikz}
\usepackage{amsmath}
\usepackage{subcaption}
\usepackage{mathpartir}
\usetikzlibrary{positioning}
\usetikzlibrary{calc}
\usetikzlibrary{decorations.pathreplacing}
\usepackage{forest}
\forestset{
dg edges/.style={for tree={parent anchor=south, child anchor=north,align=center,base=bottom,where n children=0{tier=word,edge=dotted,calign with current edge}{}}},
}

\usepackage{algpseudocode,algorithm}
\usepackage{cleveref}
\crefformat{subsection}{\S#2#1#3}

\DeclareMathOperator*{\argmax}{arg\,max}

\usepgflibrary{arrows.meta}
\usetikzlibrary{fit}
\usepackage{multirow, multicol}
\usepackage{pgfplots}

\usepackage{enumitem}
\usepackage{booktabs}       

\setlist[itemize]{noitemsep}

\DeclareDocumentCommand{\trapezoid}{O{2.0} O{1.0} O{0.5} m m}{
    \begin{scope}[scale=0.9,thick]
        \draw[anchor=mid] (0, 0) -- (0, -{#2}) node[below=9pt,anchor=base] {\ensuremath{#4}} -- ({#1}, -{#2}) node [below=9pt,anchor=base] {\ensuremath{#5}} -- ({#1}, -{#3}) -- cycle;
    \end{scope}
}

\DeclareDocumentCommand{\trapezoidd}{O{2.0} O{1.0} O{0.5} m m m}{
    \begin{scope}[scale=0.9,thick]
        \draw[anchor=mid] (0, 0) -- (0, -{#2}) node[below=9pt,anchor=base] {\ensuremath{#4}} -- ({#1}, -{#2}) node [below=9pt,anchor=base] {\ensuremath{#5}} -- ({#1}, -{#3}) -- cycle;
            \draw node[] at ( $({#2}, 0.2)!0.5!(0, 0.2)$) {\ensuremath{#6}};
    \end{scope}
}

\DeclareDocumentCommand{\trapezoiddd}{O{2.0} O{1.0} O{0.5} m m m m}{
    \begin{scope}[scale=0.9,thick]
        \draw[anchor=mid] (0, 0) -- (0, -{#2}) node[below=9pt,anchor=base] {\ensuremath{#4}} -- ({#1}, -{#2}) node [below=9pt,anchor=base] {\ensuremath{#5}} -- ({#1}, -{#3}) -- cycle;
            \draw node[] at ( $({#2}, 0.7)!0.5!(0, 0.7)$) {\ensuremath{#6}};
            \draw node[] at ( $({#2}, 0.2)!0.5!(0, 0.2)$) {\ensuremath{#7}};
    \end{scope}
}

\DeclareDocumentCommand{\trapezoiddleft}{O{2.0} O{1.0} O{0.5} m m m}{
    \begin{scope}[scale=0.9,thick]
        \draw[anchor=mid] (0, 0) -- (0, -{#2}) node[below=9pt,anchor=base] {\ensuremath{#4}} -- ({#1}, -{#2}) node [below=9pt,anchor=base] {\ensuremath{#5}} -- ({#1}, -{#3}) -- cycle;
            \node at ( $({#2}, 0.5)!0.5!(0, 0.5)$) [] {\ensuremath{#6}};
    \end{scope}
}

\DeclareDocumentCommand{\trapezoiddleftt}{O{2.0} O{1.0} O{0.5} m m m m}{
    \begin{scope}[scale=0.9,thick]
        \draw[anchor=mid] (0, 0) -- (0, -{#2}) node[below=9pt,anchor=base] {\ensuremath{#4}} -- ({#1}, -{#2}) node [below=9pt,anchor=base] {\ensuremath{#5}} -- ({#1}, -{#3}) -- cycle;
            \node at ( $({#2}, 1)!0.5!(0, 1)$) [] {\ensuremath{#6}};
                        \node at ( $({#2}, 0.5)!0.5!(0, 0.5)$) [] {\ensuremath{#7}};
    \end{scope}
}

\DeclareDocumentCommand{\righttriangle}{O{0.5} O{0.5} m m }{
    \begin{scope}[scale=0.9,thick]
        \draw[anchor=mid] (0, 0) -- (0, -{#2}) node[below=9pt,anchor=base] {\ensuremath{#3}} -- ({#1}, -{#2}) node [below right=9pt and 3pt,anchor=base] {\ensuremath{#4}} -- cycle;
    \end{scope}
}

\DeclareDocumentCommand{\righttrianglee}{O{0.5} O{0.5} m m m}{
    \begin{scope}[scale=0.9,thick]
        \draw[anchor=mid](0, 0) -- (0, -{#2}) node[below=9pt,anchor=base] {\ensuremath{#3}} -- ({#1}, -{#2}) node [below right=9pt and 3pt,anchor=base] {\ensuremath{#4}} -- cycle;
        \draw node[] at ( $({#1}, 0.2)!0.5!(0, 0.2)$) {\ensuremath{#5}};
    \end{scope}
}

\DeclareDocumentCommand{\lefttriangle}{O{0.5} O{0.5} m m}{
    \begin{scope}[scale=0.9,thick]
        \draw[anchor=mid] (0, -{#2}) node[below left=9pt and 3pt,anchor=base] {\ensuremath{#3}} -- ({#1}, -{#2}) node [below=9pt,anchor=base] {\ensuremath{#4}} -- ({#1}, 0) -- cycle;
    \end{scope}
}

\DeclareDocumentCommand{\square}{O{0.5} O{0.5} m m m}{
    \begin{scope}[scale=0.9,thick]
        \draw[anchor=mid] (0, -{#2}) node[below left=9pt and 3pt,anchor=base] {\ensuremath{#3}} -- ({#1}, -{#2}) node [below=9pt,anchor=base] {\ensuremath{#4}} -- ({#1}, 0) --  (0, 0) -- cycle;
                \draw node[] at ( $({#1}, 0.2)!0.5!(0, 0.2)$) {\ensuremath{#5}};
    \end{scope}
}

\DeclareDocumentCommand{\leftclosetriangle}{O{0.5} O{0.5} m m}{
    \begin{scope}[scale=0.9,thick]
        \draw[anchor=mid] (0, -{#2}) node[below left=9pt and 3pt,anchor=base] {\ensuremath{#3}} -- ({#1}, -{#2}) node [below=9pt,anchor=base] {\ensuremath{#4}} -- ({#1}, 0) -- cycle;
        \draw[anchor=mid] ({#1}, -{#2}-0.05) -- (0, -{#2}-0.05);
    \end{scope}
}

\DeclareDocumentCommand{\leftclosetrianglee}{O{0.5} O{0.5} m m m}{
    \begin{scope}[scale=0.9,thick]
        \draw[anchor=mid] (0, -{#2}) node[below left=9pt and 3pt,anchor=base] {\ensuremath{#3}} -- ({#1}, -{#2}) node [below=9pt,anchor=base] {\ensuremath{#4}} -- ({#1}, 0) -- cycle;
        \draw[anchor=mid] ({#1}, -{#2}-0.05) -- (0, -{#2}-0.05);
        \draw node[] at ( $({#1}, 0.2)!0.5!(0, 0.2)$) {\ensuremath{#5}};
    \end{scope}
}

\DeclareDocumentCommand{\lefttrianglee}{O{0.5} O{0.5} m m m}{
    \begin{scope}[scale=0.9,thick]
        \draw[anchor=mid] (0, -{#2}) node[below left=9pt and 3pt,anchor=base] {\ensuremath{#3}} -- ({#1}, -{#2}) node [below=9pt,anchor=base] {\ensuremath{#4}} -- ({#1}, 0) -- cycle;
        \draw node[] at ( $({#1}, 0.2)!0.5!(0, 0.2)$) {\ensuremath{#5}};
    \end{scope}
}

\DeclareDocumentCommand{\rightclosetriangle}{O{0.5} O{0.5} m m}{
    \begin{scope}[scale=0.9,thick]
        \draw[anchor=mid] (0, 0) -- (0, -{#2}) node[below=9pt,anchor=base] {\ensuremath{#3}} -- ({#1}, -{#2}) node [below right=9pt and 3pt,anchor=base] {\ensuremath{#4}} -- cycle;
        \draw[anchor=mid] ({0}, -{#2}-0.05) -- ({#1}, -{#2}-0.05);

    \end{scope}
}

\DeclareDocumentCommand{\rightclosetrianglee}{O{0.5} O{0.5} m m m}{
    \begin{scope}[scale=0.9,thick]
        \draw[anchor=mid] (0, 0) -- (0, -{#2}) node[below=9pt,anchor=base] {\ensuremath{#3}} -- ({#1}, -{#2}) node [below right=9pt and 3pt,anchor=base] {\ensuremath{#4}} -- cycle;
        \draw[anchor=mid] ({0}, -{#2}-0.05) -- ({#1}, -{#2}-0.05);
        \draw node[] at ( $({#1}, 0.2)!0.5!(0, 0.2)$) {\ensuremath{#5}};
    \end{scope}
}

\DeclareDocumentCommand{\triangle}{O{0.5} O{0.5} m m m m}{
    \begin{scope}[scale=0.9,thick]
        \draw[anchor=mid] (0, 0) node[draw, circle, black, fill, scale=0.6]{} -- (-{#1}, -{#2}) node[below=9pt,anchor=base] {\ensuremath{#3}}  --  (0, -{#2}) node[below=9pt,anchor=base] {\ensuremath{#4}} -- ({#1}, -{#2}) node [below right=9pt and 3pt,anchor=base] {\ensuremath{#5}} -- cycle;
        \draw node at ( 0, 0.5) {\ensuremath{#6}};
    \end{scope}
}

\DeclareDocumentCommand{\triangleclose}{O{0.5} O{0.5} m m m m}{
    \begin{scope}[scale=0.9,thick]
        \draw[anchor=mid] (0, 0) node[draw, circle, black, fill, scale=0.6]{} -- (-{#1}, -{#2}) node[below=9pt,anchor=base] {\ensuremath{#3}}  --  (0, -{#2}) node[below=9pt,anchor=base] {\ensuremath{#4}} -- ({#1}, -{#2}) node [below right=9pt and 3pt,anchor=base] {\ensuremath{#5}} -- cycle;
        \draw[anchor=mid] (-{#1}, -{#2}-0.05) -- ({#1}, -{#2}-0.05);
        \draw node[] at ( 0, 0.5) {\ensuremath{#6}};
    \end{scope}
}

\DeclareDocumentCommand{\triangleleft}{O{0.5} O{0.5} O{0.5} m m m m}{
    \begin{scope}[scale=0.9,thick]
        \draw[anchor=mid] (0, 0) node[draw, circle, black, fill, scale=0.6](head){} -- (-{#1}, -{#2}) node[below=9pt,anchor=base] {\ensuremath{#4}}  --  (0, -{#2}) node[below=9pt,anchor=base] {\ensuremath{#5}} -- ({#1}, -{#2}) node [below right=9pt and 3pt,anchor=base] {\ensuremath{#6}} -- cycle;
        \draw node[draw, circle, black, fill, scale=0.4] at (-{#3}, -{#2}) (node1) {};
        \draw node[below=9pt, anchor=base] at (-{#3}, -{#2}) {$h$};
        \draw[->] (node1) to[out=90] (head);
        \draw node[] at ( 0, 0.5) {\ensuremath{#7}};
        
    \end{scope}
}

\DeclareDocumentCommand{\triangleright}{O{0.5} O{0.5} O{0.5} m m m m}{
    \begin{scope}[scale=0.9,thick]
        \draw[anchor=mid] (0, 0) node[draw, circle, black, fill, scale=0.6](head){} -- (-{#1}, -{#2}) node[below=9pt,anchor=base] {\ensuremath{#4}}  --  (0, -{#2}) node[below=9pt,anchor=base] {\ensuremath{#5}} -- ({#1}, -{#2}) node [below right=9pt and 3pt,anchor=base] {\ensuremath{#6}} -- cycle;
        \draw node[draw, circle, black, fill, scale=0.4] at ({#3}, -{#2}) (node1) {};
        \draw node[below=9pt, anchor=base] at ({#3}, -{#2}) {$h$};
        \draw[->] (node1) to[in=45,out=90] (0,0);
        \draw node[] at ( 0, 0.5) {\ensuremath{#7}};
        
    \end{scope}
}

\title{Headed-Span-Based Projective Dependency Parsing}

\author{Songlin Yang, Kewei Tu\thanks{\; Corresponding Author}\\
  School of Information Science and Technology, ShanghaiTech University \\
    Shanghai Engineering Research Center of Intelligent Vision and Imaging\\ 
    {\tt \{yangsl,tukw\}@shanghaitech.edu.cn}\\
 }

\begin{document}
\maketitle
\begin{abstract}
 We propose a new method for projective dependency parsing based on headed spans. 
In a projective dependency tree, the largest subtree rooted at each word covers a contiguous sequence (i.e., a span) in the surface order. We call such a span marked by a root word \textit{headed span}.
 A projective dependency tree can be represented as a collection of headed spans. We decompose the score of a dependency tree into the scores of the headed spans and design a novel $O(n^3)$ dynamic programming algorithm to enable global training and exact inference.  Our model achieves state-of-the-art or competitive results on PTB, CTB, and UD \footnote{Our code is publicly available at \url{https://github.com/sustcsonglin/span-based-dependency-parsing}}. 




\end{abstract}

\section{Introduction}

Dependency parsing is an important task in natural language processing, which has numerous applications in downstream tasks, such as opinion mining \cite{zhang-etal-2020-syntax}, relation extraction \cite{DBLP:conf/aaai/JinS0XM020},  named entity recognition \cite{jie-lu-2019-dependency}, machine translation \cite{bugliarello-okazaki-2020-enhancing}, among others. 

There are two main paradigms in dependency parsing: graph-based and transition-based methods. Graph-based methods decompose the score of a tree into the scores of parts. In the simplest first-order graph-based methods \cite[][\it{inter alia}]{mcdonald-etal-2005-online}, the parts are single dependency arcs. In higher-order graph-based methods \cite{mcdonald-pereira-2006-online, carreras-2007-experiments, koo-collins-2010-efficient, ma-zhao-2012-fourth},  the parts are combinations of multiple arcs. Transition-based methods \cite[][\it{inter alia}]{nivre-scholz-2004-deterministic, chen-manning-2014-fast} read the sentence sequentially and conduct a series of local decisions to build the final parse. 
Recently, transition-based methods with Pointer Networks \cite{DBLP:conf/nips/VinyalsFJ15} have obtained competitive performance to graph-based methods \cite{ma-etal-2018-stack, liu-etal-2019-hierarchical, fernandez-gonzalez-gomez-rodriguez-2019-left, DBLP:journals/corr/abs-2105-09611}. 
 
 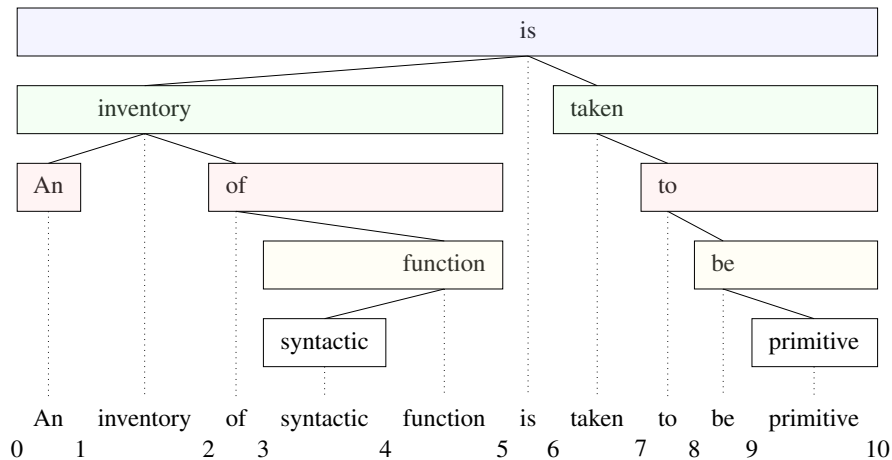
\begin{figure*}[tb!]
\centering
\scalebox{.85}{
\begin{forest}
dg edges
[is, name=top6, 
  [inventory, name=top2
   [An, name=top1 [  An, name=node1]]
   [inventory, name=node2] 
   [of, name=top3[ of, name=node3]
   [function,name=top5
   [syntactic,name=top4 [syntactic, name=node4]]
   [function, name=node5]]
   ]
   ]
   [is,name=node6]
   [taken,name=top7[taken, name=node7]
    [to,name=top8[to, name=node8]
    [be,name=top9 [be, name=node9]
    [primitive,name=top10 [primitive,name=node10]
    ]
    ]
    ]
    ]
]
\foreach \i in {2, ..., 10}{
    \pgfmathtruncatemacro{\j}{\i-1}
\node (boundary\j) [below =1ex of node1] at  ( $ (node\j.east)!0.5!(node\i.west) $ )   {\j};
}
\node (boundary0) [left=3ex of boundary1] {0}[];
\node (boundary10) [right=8ex of boundary9] {10}[];
\node (bos) [left=2.5ex of node1.center] {}[];
\node (eos) [right=2ex of node10] {}[];
\node[draw,fit=(top6)(boundary0.center|-top6.center)(boundary10.center|-top6.center),inner sep=0pt,fill=blue!20,fill opacity=0.2] (level11) {};
\node[draw,fit=(top2)(boundary0.center|-top2.center)(boundary5.center|-top2.center),inner sep=0pt, fill=green!20, fill opacity=0.2] (level21) {};
\node[draw,fit=(top7)(boundary6.center|-top7.center)(boundary10.center|-top7.center),inner sep=0pt, fill=green!20, fill opacity=0.2] (level22) {};
\node[draw,fit=(top1)(boundary0.center|-top1.center)(boundary1.center|-top1.center),inner sep=0pt, fill=red!20, fill opacity=0.2] (level31) {};
\node[draw,fit=(top3)(boundary2.center|-top3.center)(boundary5.center|-top3.center),inner sep=0pt, fill=red!20, fill opacity=0.2] (level32) {};
\node[draw,fit=(top8)(boundary7.center|-top8.center)(boundary10.center|-top8.center),inner sep=0pt, fill=red!20, fill opacity=0.2] (level33) {};
\node[draw,fit=(top5)(boundary3.center|-top5.center)(boundary5.center|-top5.center),inner sep=0pt, fill=yellow!20, fill opacity=0.2] (level41) {};
\node[draw,fit=(top9)(boundary8.center|-top9.center)(boundary10.center|-top9.center),inner sep=0pt, fill=yellow!20, fill opacity=0.2] (level42) {};
\node[draw,fit=(top4)(boundary3.center|-top4.center)(boundary4.center|-top4.center),inner sep=0pt, ] (level51) {};
\node[draw,fit=(top10)(boundary9.center|-top10.center)(boundary10.center|-top10.center),inner sep=0pt] (level52) {};
\end{forest}
}
\caption{Illustration of a projective dependency parse tree. Each rectangle represents a headed span. }
\label{tree_example}
\end{figure*}

A main limitation of first-order graph-based methods is that they independently score each arc based solely on the two words connected by the arc. Ideally, the appropriateness of an arc should depend on the whole parse tree, particularly the subtrees rooted at the two words connected by the arc. Although subtree 
information could be implicitly encoded \cite{falenska-kuhn-2019-non} in powerful neural encoders such as LSTMs \cite{hochreiter1997long} and Transformers \cite{DBLP:conf/nips/VaswaniSPUJGKP17}, 
there is evidence that their encoding of such information is inadequate.  For example, higher-order graph-based methods, which capture more subtree information by simultaneously considering multiple arcs, have been found to outperform first-order methods despite using powerful encoders \cite{fonseca-martins-2020-revisiting,zhang-etal-2020-efficient,wang-tu-2020-second}. In contrast to the line of work on higher-order parsing, we propose a different way to incorporate more subtree information as discussed later. 


Transition-based methods, on the other hand, can easily utilize information from partially built subtrees, but they have their own shortcomings. For instance, most of them cannot perform global optimization during decoding \footnote{We are aware of few transition-based parsers performing global optimization via dynamic programming algorithms, cf. \citet{kuhlmann-etal-2011-dynamic, shi-etal-2017-fast, gomez-rodriguez-etal-2018-global}.} and rely on greedy or beam search to find a locally optimal parse,
and their sequential decoding may cause error propagation as past decision mistakes will negatively affect the decisions in the future.

To overcome the aforementioned limitations of first-order graph-based and transition-based methods, we propose a new method for projective dependency parsing based on so-called headed spans.
  A projective dependency tree has a nice structural property that the \textbf{largest} subtree rooted at each word covers a contiguous sequence (i.e., a span) in the surface order. We call such a span marked with its root word a \textit{headed span}.
  A projective dependency tree can be treated as a collection of headed spans such that each word corresponds to exactly one headed span, as illustrated in Figure \ref{tree_example}. For example, $(0, 5, \text{inventory})$ is a headed span, in which span $(0, 5)$ has a head word $\textit{inventory}$.
   In this view, projective dependency parsing is similar to constituency parsing as a constituency tree can be treated as a collection of constituent spans. The main difference is that in a binary constituency tree, a constituent span $(i, k)$ is made up by two adjacent spans $(i, j)$ and $(j, k)$, while in a projective dependency tree, a headed span $(i, k, x_h)$ is made up by one or more smaller headed spans and a single word span $(h-1, h)$. For instance, $(0, 5, \text{inventory})$ is made up by $(0, 1, \text{An}), (1, 2)$ and $(2, 5, \text{of})$. There are a few constraints between headed spans to force projectivity (\cref{sec:parsing}). These structural constraints are the key to designing an efficient dynamic programming algorithm for exact inference. 
  
   Because of the similarity between constituency parsing and our head-span-based view of projective dependency parsing, we can draw inspirations from the constituency parsing literature to design our dependency parsing method.
   Specifically, span-based constituency parsers \cite{stern-etal-2017-minimal, kitaev-klein-2018-constituency, DBLP:conf/ijcai/ZhangZL20, xin-etal-2021-n} decompose the score of a constituency tree into the scores of its constituent spans and use the CYK algorithm \cite{Cocke1969ProgrammingLA, Younger1967RecognitionAP, Kasami1965AnER} for global training and inference. Built upon powerful neural encoders, they have obtained state-of-the-art performance in constituency parsing. Inspired by them, we propose to decompose the score of a projective dependency tree into the scores of headed spans and design a novel $O(n^3)$ dynamic programming algorithm for global training and exact inference, which is on par with the Eisner algorithm \cite{eisner-1996-three} in time complexity for projective dependency parsing. 
        We make a departure from existing graph-based methods since we do not model dependency arcs directly. Instead, the dependency arcs are \textit{induced} from the collection of headed spans (\cref{sec:parsing}). Compared with first-order graph-based methods, our method can utilize more subtree information since a headed span contains all children (if any) of the corresponding headword (and all words within the subtree). 
  Compared with most of transition-based methods, our method allows global training and exact inference and does not suffer from error propagation or exposure bias.

   Our contributions can be summarized as follows:

\begin{figure*}[tb!]
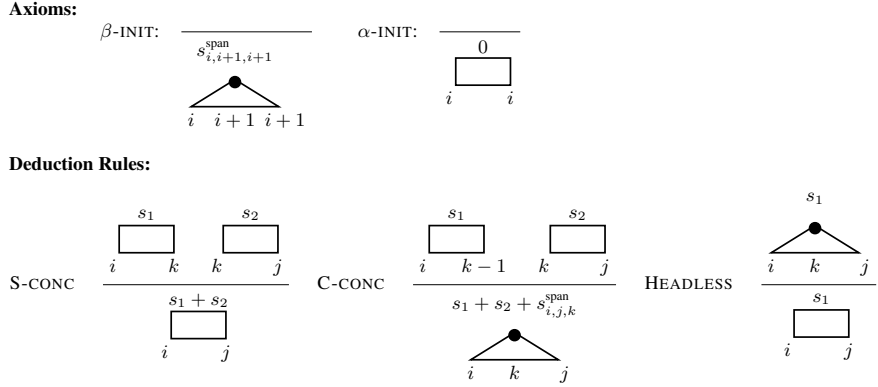

    \begin{subfigure}{\textwidth}
        \centering
        \small 
        \scalebox{.8}{
        \begin{tabular}{rlrlrl}
            \multicolumn{2}{l}{\textbf{Axioms:}}
            &
            \\
            \multicolumn{4}{c}{
            \textsc{$\beta$-init}:\quad
            $\inferrule{ }{
        \tikz[baseline=-15pt]{\triangle[0.8][0.5]{i}{i+1}{i+1}{s^{\text{span}}_{i, i+1, i+1}}}
            }$
            \quad\quad
            \textsc{$\alpha$-init}:\quad
            $\inferrule{ }{
                \tikz[baseline=-10pt]{\square[1][0.5]{i}{i}{0}}
            }$
            }
            \vspace{10pt}
            \\
            \multicolumn{2}{l}{\textbf{Deduction Rules:}}
            &
            \vspace{5pt}
            \\
		          \textsc{S-conc}
        &
                        $\inferrule{
          \tikz[baseline=-9pt]{\square[1][0.5]{i}{k}{s_1}} \quad 
             \tikz[baseline=-9pt]{\square[1][0.5]{k}{j}{s_2}} 
            } {
                \tikz[baseline=-6pt]{\square[1][0.5]{i}{j}{s_1+s_2}}
            }$
            &
        \textsc{C-conc}
        &
         $\inferrule{
          \tikz[baseline=-9pt]{\square[1][0.5]{i}{k-1}{s_1}} \quad 
             \tikz[baseline=-9pt]{\square[1][0.5]{k}{j}{s_2}} 
            } {
                \tikz[baseline=+3pt]{\triangle[0.8][0.5]{i}{k}{j}{s_1+s_2+s^{\text{span}}_{i, j, k}}}
            }$
        &
        \textsc{Headless}
        &
                 $\inferrule{
                \tikz[baseline=-15pt]{\triangle[0.8][0.5]{i}{k}{j}{s_1}}
            } {
                \tikz[baseline=-6pt]{\square[1][0.5]{i}{j}{s_1}}
            }$
        
	      \end{tabular}
	      }
	    \end{subfigure}
    \caption{
    Deduction rules for our proposed parsing algorithm. All deduction items are annotated with their scores.
    }
    \label{fig:deduction}

\end{figure*}

 \begin{itemize}
 	\item We treat a projective dependency tree as a collection of headed spans, providing a new perspective of projective dependency parsing. 
 	\item We design a novel $O(n^3)$ dynamic programming algorithm to enable global training and exact inference for our proposed model. 
 	\item We have obtained the state-of-the-art or competitive results on PTB, CTB, and UD v2.2, showing the effectiveness of our proposed method.  
 \end{itemize}

\section{Parsing}












We adopt the two-stage parsing strategy, i.e., we first predict an unlabeled tree and then predict the dependency labels. Given a sentence $x_1, ..., x_n$, its unlabeled projective dependency parse tree $y$ can be regarded as a collection of headed spans $(l_i, r_i, x_i)$ where $1\le i \le n$. For each word $x_i$, we can find exactly one headed span $(l_i, r_i, i)$ (where $l_i$ and $r_i$ are the left and right span boundaries) given parse tree $y$, so there are totally $n$ headed spans in $y$ as we can see in Figure \ref{tab:ptb_ctb}.  We can use a simple post-order traversal algorithm to obtain all headed spans in $O(n)$ time.  We then define the score of $y$ as:
\[
s(y) = \sum_{i=1,...,n} s_{l_i,r_i,i}^{\text{span}}
\]
and we show how to compute them using neural networks in the next section. 

Our parsing algorithm is based on the following key observations:
\begin{itemize} 
	\item For a given parent word $x_k$, if it has any children to the left (right), then all headed spans of its children in this direction should be consecutive and form a larger span, which we refer to as the left (right) child span. The left (right) boundary of the headed span of $x_k$ is the left (right) boundary of the leftmost (rightmost) child span, or $k-1$ ($k$) if $x_k$ has no child to the left (right). 
	\item  If a parent word $x_k$ has children in both directions, then its left span and right span are separated by the single word span $(k-1, k)$. 
	\end{itemize}
	
Based on these observations, we design the following parsing items: (1)
 $\alpha_{i, j}$: the accumulated score of span $(i, j)$ serving as a left or right child span. (2) $\beta_{i, j, k}$: the accumulated score of the headed span $(i, j, k)$.
We use the parsing-as-deduction framework \cite{pereira-warren-1983-parsing} to describe our algorithm in Fig. \ref{fig:deduction}. We draw $\alpha_{i, j}$ as rectangles and $\beta_{i, j, k}$ as triangles.  The rule $\textsc{S-conc}$ is used to concatenate two consecutive child spans into a single child span; $\textsc{C-conc}$ is used to concatenate left and right child span $(i, k-1)$ and $(k, j)$ along with the root word-span ($k-1, k)$ to form a headed span $(i, j, k)$; $\textsc{Headless}$ is used to obtain a headless child span from a headed span.
Fig. \ref{fig:deduction} corresponds to the following recursive formulas:
\begin{align}
\beta_{i, i+1, i+1} &= s_{i, i+1, i+1}^{\text{span}} \label{dp1} \\
\alpha_{i, i} &= 0 \\ 
\beta_{i, j, k} &=  \alpha_{i, k-1} + \alpha_{k, j} + s_{i, j, k}^{\text{span}} \label{dp3} \\
\alpha_{i, j} &= \max ( \max\limits_{i <k < j}(\alpha_{i, k} + \alpha_{k, j}), \nonumber \\ & \phantom{\alpha(i, i+1)}\max\limits_{i<h\le j} (\beta_{i, j, h})) \label{dp2}
\end{align}
We set $\alpha_{i, i} = 0$ for the convenience of calculating $\beta_{i,j,k}$ when $x_k$ does not have children on either side.
 In Eq. \ref{dp2}, we can see that the child span comes from either multiple smaller consecutive child spans (i.e., $ \max\limits_{i <k < j}(\alpha(i, k) + \alpha(k, j))$) or a single headed span (i.e., $\max\limits_{i<h\le j} (\beta(i, j, h)))$). We also maintain backpointers based on these equations (i.e., maintain all $\argmax$) for parsing.
 
 \begin{figure*}[!ht]
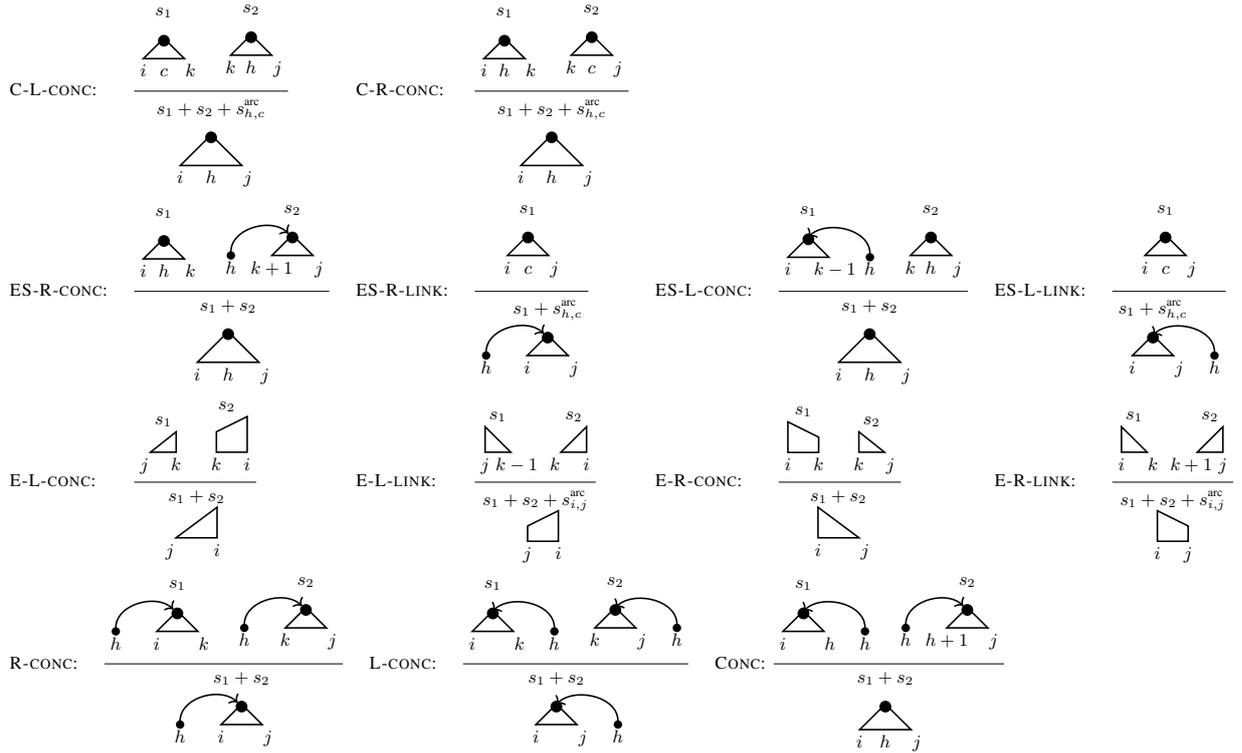

    \centering
        \small 
\begin{subfigure}{\linewidth}
            \scalebox{0.75}{
            \begin{tabular}{llllllllll}        
\textsc{C-L-conc}:
    &
        $\inferrule{\tikz[]{\triangle[0.4][0.4]{i}{c}{k}{s_1}} \quad
    \tikz[]{\triangle[0.4][0.4]{k}{h}{j}{s_2}}} {
    \tikz[]{\triangle[0.6][0.6]{i}{h}{j}{s_1 + s_2 + s^{\text{arc}}_{h, c}}}
    }$
    &
        \textsc{C-R-conc}:
    &
            $\inferrule{\tikz[]{\triangle[0.4][0.4]{i}{h}{k}{s_1}} \quad
    \tikz[]{\triangle[0.4][0.4]{k}{c}{j}{s_2}}} {
    \tikz[]{\triangle[0.6][0.6]{i}{h}{j}{s_1 + s_2 + s^{\text{arc}}_{h, c}}}
    }$
\\
    \textsc{ES-R-conc}:
    &
        $\inferrule{\tikz[]{\triangle[0.4][0.4]{i}{h}{k}{s_1}} \quad
    \tikz[]{\triangleleft[0.4][0.4][1.2]{k+1}{}{j}{s_2}}   } {
    \tikz[]{\triangle[0.6][0.6]{i}{h}{j}{s_1 + s_2}}
    }$
    &
    \textsc{ES-R-link}:
    &
        $\inferrule{
    \tikz[]{\triangle[0.4][0.4]{i}{c}{j}{s_1}}} {
    \tikz[]{\triangleleft[0.4][0.4][1.2]{i}{}{j}{s_1 + s^{\text{arc}}_{h,c}}
    }}$
    &
    	\textsc{ES-L-conc}:
	&
	    $\inferrule{\tikz[]{\triangleright[0.4][0.4][1.2]{i}{}{k-1}{s_1}} \quad
    \tikz[]{\triangle[0.4][0.4]{k}{h}{j}{s_2}}   } {
    \tikz[]{\triangle[0.6][0.6]{i}{h}{j}{s_1 + s_2}}
    }$
    &
    \textsc{ES-L-link}:
    &
     $\inferrule{
    \tikz[]{\triangle[0.4][0.4]{i}{c}{j}{s_1}} }
     {
    \tikz[]{\triangleright[0.4][0.4][1.2]{i}{}{j}{s_1 + s_{h,c}^{\text{arc}}}
    }}$    
%
\\
            \textsc{E-L-conc}:
            &
            $\inferrule{
                \tikz[baseline=-9pt]{\lefttrianglee[0.5][0.4]{j}{k}{s_1}}\quad
                \tikz[baseline=-9pt]{\trapezoiddleft[0.6][0.4][-0.3]{k}{i}{s_2}}
            }{
                \tikz[baseline=-10pt]{\lefttrianglee[0.8][0.6]{j}{i}{s_1+s_2}}
            }$
            &
            \textsc{E-L-link}:
            &
            $\inferrule{
                \tikz[baseline=-9pt]{\righttrianglee[0.5][0.5]{j}{k-1}{s_1}}
                \tikz[baseline=-9pt]{\lefttrianglee[0.5][0.5]{k}{i}{s_2}}
            } {
                \tikz[baseline=-6pt]{\trapezoiddleft[0.6][0.3][-0.3]{j}{i}{s_1+s_2+s^{\text{arc}}_{i,j}}}
            }$
            &
                        \textsc{E-R-conc}:
            &
            $\inferrule{
                \tikz[baseline=-15pt]{\trapezoidd[0.6][0.6][0.3]{i}{k}{s_1}} \quad
                \tikz[baseline=-10pt]{\righttrianglee[0.5][0.4]{k}{j}{s_2}
                }
            } {
                \tikz[baseline=-10pt]{\righttrianglee[0.8][0.6]{i}{j}{s_1+s_2}}
            }$
            &
            \textsc{E-R-link}:
            &
            $\inferrule{
                \tikz[baseline=-9pt]{\righttrianglee[0.5][0.5]{i}{k}{s_1}}
                \tikz[baseline=-9pt]{\lefttrianglee[0.5][0.5]{k+1}{j}{s_2}}
            } {
                \tikz[baseline=-10pt]{\trapezoidd[0.6][0.6][0.3]{i}{j}{s_1+s_2+s^{\text{arc}}_{i, j}}}
            }$

    \end{tabular}
    }
\end{subfigure}
\begin{subfigure}{\linewidth}
            \scalebox{0.75}{
            \begin{tabular}{llllllll}        

    \textsc{R-conc}:
    &
        $\inferrule{\tikz[]{\triangleleft[0.4][0.4][1.2]{i}{}{k}{s_1}} \quad
    \tikz[]{\triangleleft[0.4][0.4][1.2]{k}{}{j}{s_2}}} {
    \tikz[]{\triangleleft[0.4][0.4][1.2]{i}{}{j}{s_1+s_2}}
    }$
    &
   \textsc{L-conc}:
    &
        $\inferrule{\tikz[]{\triangleright[0.4][0.4][1.2]{i}{}{k}{s_1}} \quad
    \tikz[]{\triangleright[0.4][0.4][1.2]{k}{}{j}{s_2}}} {
    \tikz[]{\triangleright[0.4][0.4][1.2]{i}{}{j}{s_1+s_2}}
    }$
    &
    \textsc{Conc}:
    $\inferrule{\tikz[]{\triangleright[0.4][0.4][1.2]{i}{}{h}{s_1}} \quad
    \tikz[]{\triangleleft[0.4][0.4][1.2]{h+1}{}{j}{s_2}}} {
    \tikz[]{\triangle[0.5][0.5]{i}{h}{j}{s_1+s_2}}
    }$
    &
    \end{tabular}
\\
}
\end{subfigure}
\caption{Deductive rules of the parsing algorithms of \citet{collins-1996-new} (the first line), \citet{eisner-satta-1999-efficient} (the second line), \citet{eisner-1997-bilexical} (the third line). The last line is the resulting deduction rules after applying head-splitting on $\textsc{ES-L-conc}$ and $\textsc{ES-R-conc}$. All deduction items are annotated with their scores. We only consider the pure dependency versions of these algorithms. We omit axiom items for simplicity.
}
\label{fig:compare}
\end{figure*}

A key point of our parsing algorithm is that, during backtracking, we add arcs emanated from the headword of a large headed span to \textbf{every} headword of (zero or more) smaller headed spans within the left/right child span, so that we can induce a dependency tree. Finding all smaller headed spans within left and right child spans requires finding the best segmentation, which is similar to the inference procedure of the semi-Markov CRF model \cite{DBLP:conf/nips/SarawagiC04}. We provide the pseudocode of our parsing algorithm in Appd. A.

\paragraph{Parsing complexity.} From Eq. \ref{dp1} to \ref{dp2}, we can see that at most three variables (i.e., $i,j,k$) are required to iterate over and therefore the total parsing time complexity is $O(n^3)$. 
\paragraph{Spurious ambiguity.} Note that different order of concatenation of child spans can result in the same parse, although this does not affect finding the optimal parse. 
\paragraph{Comparison with previous parsing algorithms.} 
We compare our algorithm with three classical parsing algorithms \citep{collins-1996-new,eisner-satta-1999-efficient,eisner-1997-bilexical} in order to help readers better understand our algorithm. We only consider their pure dependency versions\footnote{The parsing algorithms of \citet{collins-1996-new} and \citet{eisner-satta-1999-efficient} are defined with (lexicalized) context-free gramars. \citet{gomez-rodriguez-etal-2008-deductive, gomez-rodriguez-etal-2011-dependency} provide their pure dependency versions, which amounts to considering arc scores only.} for the convenience of discussion. Fig. \ref{fig:deduction} shows the deductive rules of the three algorithms.

\citet{collins-1996-new} adapt the CYK algorithm by maintaining head positions for both sides, thereby increasing the parsing complexity from $O(n^3)$ to $O(n^5)$.
Their parsing items are identified by two endpoints and a head position, which is similar to our concept of headed spans superficially. However, 
in their algorithm, there could be multiple spans sharing the same head position within a single parse. For instance, $(i, j)$ and $(k, j)$ share the same head position $h$ in \textsc{C-L-conc}. In contrast, spans cannot share a head position in a single parse under our definition, because there is exactly one headed span for each word.  Besides, the concatenation order of subtrees differs.

\citet{eisner-satta-1999-efficient} note that the linking of heads and the concatenation of subtrees can be separated (e.g., $\textsc{C-R-conc}$ can be decomposed into two rules, $\textsc{ES-R-conc}$ and $\textsc{ES-R-link}$)
so that the parsing complexity can be reduced to $O(n^4)$. This strategy is also known as the hook trick, which reduces subtrees to headless spans (e.g., $(i, c, j)$ to $(i, j)$ in  \textsc{ES-l-link} and \textsc{ES-r-link}).

\citet{eisner-1997-bilexical} uses the head-splitting trick to decrease parsing complexity to $O(n^3)$. The key idea is to split each subtree into a left and a right fragment, so that the head is always placed at one of the two boundaries of a fragment instead of an internal position, thereby eliminating the need of maintaining the head positions.


Our algorithm adopts a combination of the hook trick and the head-splitting trick. Starting from the rules of \citet{eisner-satta-1999-efficient} that apply the hook trick, we can rewrite \textsc{ES-L-conc}, \textsc{ES-R-conc} as \textsc{L-conc}, \textsc{R-conc} and \textsc{Conc}. It is easy to verify the equivalence of the rules before and after the rewrite\footnote{Note that this only holds for the pure dependency version, since otherwise we cannot track some intermediate constituent spans after changing the concatenation order of subtrees.}. The key difference is in the concatenation order of subtrees. We concatenate all subtrees to the left/right of the new head first, which can be viewed as adopting the head-splitting trick. Then, note that the position of the head is uniquely determined by the two concatenations of subtrees, and that our model does not consider $s^{\text{arc}}$. Consequently, we have no need to maintain head position $h$ in \textsc{L-conc} and \textsc{R-conc} and can merge these two rules to \textsc{S-conc} of \cref{fig:deduction}. Accordingly, \textsc{Conc} can be modified to \textsc{C-conc} of \cref{fig:deduction}. Eliminating bookkeeping of $h$ is how we can obtain better parsing complexity than \citet{eisner-satta-1999-efficient}. Finally, we can incorporate span score $s^{\text{span}}_{i,j,h}$ into \textsc{C-conc}. 
\section{Model}
\label{sec:parsing}

\subsection{Neural encoding and scoring}
\label{sec:neural}
We add <bos> (beginning of sentence) at $x_0$ and <eos> (end of sentence) at $x_{n+1}$. In the embedding layer, we apply mean-pooling to the last layer of BERT \cite{devlin-etal-2019-bert} (i.e., taking the mean value of all subword embeddings) to generate dense word-level representation $e_i$ for each token $x_i$ \footnote{For some datasets (e.g., Chinese Treebank), we concatenate the POS tag embedding with the BERT embedding as $e_i$.}. Then we feed $e_0, ..., e_{n+1}$ into a 3-layer bidirectional LSTM (BiLSTM) to get $c_0, ..., c_{n+1}$, where $c_i = [f_i; b_i]$ and $f_i$ and $b_i$ are the forward and backward hidden states of the last BiLSTM layer at position $i$ respectively. We then use the fencepost representation, which is commonly used in constituency parsing \cite{cross-huang-2016-span, stern-etal-2017-minimal}, to encode span $(i, j)$ as $e_{i,j}$:
\begin{align*}
 h_{k}=\left[f_{k}, b_{k+1}\right] \\
 e_{i, j} = h_{j} - h_{i}	 
\end{align*}

After obtaining the word and span representations, we use deep biaffine function \cite{DBLP:conf/iclr/DozatM17} to score headed spans:
\begin{align*}
	c^{\prime}_k &= \text{MLP}_{\text{word}}(c_k) \\ 
	e^{\prime}_{i, j} &= \text{MLP}_{\text{span}}(e_{i,j}) \\
	s_{i, j, k}^{\text{span}} &=\left[c_{k}^{\prime} ; 1\right]^{\top}W^{\text{span}}\left[e_{i,j}^{\prime} ; 1\right]
\end{align*}
where $\text{MLP}_{\text{word}}$ and $\text{MLP}_{\text{span}}$ are multi-layer perceptrons (MLPs) that project word and span representations into $d$-dimensional spaces respectively; $W^{\text{span}} \in \mathcal{R}^{(d+1) \times (d+1)}$. 

Similarly, we use deep biaffine functions to score the labels of dependency arcs for a given gold or predicted tree \footnote{ In our preliminary experiments, we find that directly calculating the scores based on parent-child word representations leads to a slightly better result (< 0.1 LAS) than those based on span representations. A possible reason is that, since LAS is arc-factorized, even if we predict a correct parent-child pair, we can predict the wrong headed spans for the parent or child or both, thereby negatively affecting the labeling scores and resulting in worse LAS. Therefore, in our work we use arc-based label scores to suit the LAS metric. }:
\begin{align*}
	c^{\prime}_{i} &= \text{MLP}_{\text{parent}} (c_{i}) \\ 
	c^{\prime}_{j} &= \text{MLP}_{\text{child}} (c_{j}) \\
	s_{i,j,r}^{\text{label}} &= \left[c_{i}^{\prime} ; 1\right]^{\top}W^{\text{label}}_{r}\left[c_{j}^{\prime} ; 1\right]
\end{align*}
 \noindent
where $\text{MLP}_{\text{parent}}$ and $\text{MLP}_{\text{child}}$ are MLPs that map word representations into $d^{\prime}$-dimensional spaces; $W^{\text{label}}_{r} \in \mathcal{R}^{(d^{\prime}+1) \times (d^{\prime}+1)}$ for each relation type $r \in R$ in which $R$ is the set of all relation types.

\subsection{Training loss}
\label{sec:loss}
Following previous work, we decompose the training loss into the unlabeled parse loss and  arc label loss:
\[
L = L_{\text{parse}} + L_{\text{label}}
\]

For $L_{\text{parse}}$, we can either design a local span-selection loss:
\[
L^{\text{local}}_{\text{parse}} = \sum_{(i, j, k) \in y} -\log  \frac{\exp(s^{\text{span}}_{i,j,k})}{\sum\limits_{0 \le p \le k < q \le n} \exp(s^{\text{span}}_{p,q,k})} 
\]
which is akin to the head-selection loss \cite{DBLP:conf/iclr/DozatM17}, or use global structural loss. Experimentally, we find that the max-margin loss \cite{taskar-etal-2004-max} (also known as structured SVM) performs best. The max-margin loss aims to maximize the margin between the score of the gold tree $y$ and the incorrect tree $y^{\prime}$ of the highest score:
\begin{equation}
  L_{\text{parse}} = \max (0, \max _{y^{\prime} \neq y}(s(y^{\prime}) + \Delta(y^{\prime}, y) - s(y))  
\label{mm_loss}
\end{equation} \noindent
where $\Delta$ measures the difference between the incorrect tree and gold tree. Here we let $\Delta$ to be the Hamming distance (i.e., the total number of mismatches of headed spans).  We can perform cost-augmented inference \cite{DBLP:conf/icml/TaskarCKG05} to compute Eq. \ref{mm_loss}.

%
Finally, we use cross entropy for $L_{\text{label}}$:
\[
L_{\text{label}} =  \sum_{ (x_i \rightarrow x_j, r) \in y} -\log \frac{\exp(s_{i, j, r}^{\text{label}})}{\sum\limits_{r^{\prime} \in R}\exp(s^{\text{label}}_{i, j, r^{\prime}}) }
\]
where 
$(x_i \rightarrow x_j, r) \in y$ denotes every dependency arc from $x_i$ to $x_j$ with label $r$ in $y$.


\section{Experiments}
\subsection{Data and setting}
Following \citet{wang-tu-2020-second}, we evaluate our proposed method on Penn Treebank (PTB) 3.0 \cite{marcus-etal-1993-building}, Chinese Treebank (CTB) 5.1 \cite{DBLP:journals/nle/XueXCP05} and 12 languages on Universal Dependencies (UD) 2.2: BG-btb, CA-ancora, CS-pdt, DE-gsd, EN-ewt, ES-ancora, FR-gsd, IT-isdt, NL-alpino, NO-rrt, RO-rrt, RU-syntagrus \footnote{We do not concatenate all datasets during training. We train on each dataset separately.}.
For PTB, we use the Stanford Dependencies conversion software of version 3.3 to obtain dependency trees. For CTB, we use head-rules from \citet{zhang-clark-2008-tale} and Penn2Malt\footnote{\url{https://cl.lingfil.uu.se/~nivre/research/Penn2Malt.html}} to 
obtain dependency trees.
   Following \citet{wang-tu-2020-second}, we use gold POS tags for CTB and UD. We do not use POS tags in PTB. 
 For PTB/CTB, we drop all nonprojective trees during training. For UD, we use MaltParser v1.9.2 \footnote{\url{http://www.maltparser.org/download.html}} to adopt the pseudo-projective transformation \cite{nivre-nilsson-2005-pseudo} to convert nonprojective trees into projective trees when training, and convert back when evaluating, for both our model and reimplemented baseline model. See Appd. B for implementation details.
 
\subsection{Evaluation methods}
We report the unlabeled attachment score (UAS) and labeled attachment score (LAS) averaged from three runs with different random seeds. In each run, we select the model based on the performance on the development set. Following \citet{wang-tu-2020-second}, we ignore all punctuation marks during evaluation.


\begin{table}[tb!]
    \centering 
    \scalebox{.9}{
    \begin{tabular}{lcccc}
        \toprule 
        & \multicolumn{2}{c}{{\bf PTB}} & \multicolumn{2}{c}{{\bf CTB}}\\
        & UAS & LAS & UAS & LAS \\
        \midrule 
       {\it MFVI2O} & 95.98 & 94.34 & 90.81 & 89.57 \\
       {\it TreeCRF2O} & 96.14 & 94.49 & - & - \\
       {\it HierPtr} & 96.18 & 94.59 & 90.76 & 89.67 \\
       \hline 
       &\multicolumn{2}{c}{\underline{+$\text{BERT}_{\text{base}}$}}&\multicolumn{2}{c}{\underline{+$\text{BERT}_{\text{base}}$}} \\
       {\it RNGTr} & 96.66 & 95.01 & 92.98 & 91.18 \\
       \hline 
        &\multicolumn{2}{c}{\underline{+$\text{BERT}_{\text{large}}$}}&\multicolumn{2}{c}{\underline{+$\text{BERT}_{\text{base}}$}}\\
       {\it MFVI2O} & 96.91 & 95.34 & 92.55 & 91.69 \\
       {\it HierPtr} & 97.01 & 95.48 & 92.65 & 91.47 \\ 
       $\textit{Biaffine+MM}^{\dagger}$ & 97.22 & 95.71 & 93.18 & 92.10  \\
       {\it Ours} & \textbf{97.24} & \textbf{95.73} & \textbf{93.33} & \textbf{92.30} \\ 
       \hline 
        &\multicolumn{2}{c}{For reference}\\
               \hline 
        &\multicolumn{2}{c}{\underline{+$\text{XLNet}_{\text{large}}$}}&\multicolumn{2}{c}{\underline{+$\text{BERT}_{\text{base}}$}}\\
        {\it HPSG$^\flat$} & 97.20 & 95.72 & - & - \\
        {\it HPSG+LAL$^\flat$} & 97.42 & 96.26 & 94.56 & 89.28 \\
       \bottomrule 
    \end{tabular}
    }
    \caption{Results for different model on PTB and CTB. $^\flat$ indicate that they use additional annotated constituency trees in training. $^{\dagger}$ means our reimplementation. {\it Biaffine}: \citet{DBLP:conf/iclr/DozatM17}. {\it MFVI2O}: \citet{wang-tu-2020-second}. {\it TreeCRF2O}:  \citet{zhang-etal-2020-efficient}. {\it RNGTr}:   \citet{DBLP:journals/tacl/MohammadshahiH21}. {\it HierPtr}: \citet{DBLP:journals/corr/abs-2105-09611}.  {\it HPSG}: \citet{zhou-zhao-2019-head}. {\it HPSG+LAL}:   \citet{mrini-etal-2020-rethinking}.}
    \label{tab:ptb_ctb}
\end{table}

\begin{table*}[tb!]\small
	\centering
	\vskip -.0in
	{\setlength{\tabcolsep}{.8em}
		\makebox[\linewidth]{\resizebox{\linewidth}{!}{%
				\begin{tabular}{lcccccccccccc|l}
					\toprule
					  &	bg &	 ca & cs& de& en& es& fr & it&nl&no&ro&ru &Avg\\
					 \toprule
					{\it TreeCRF2O} & 90.77 & 91.29 & 91.54 & 80.46 & 87.32 & 90.86 & 87.96 & 91.91 & 88.62 & 91.02 & 86.90 & 93.33 & 89.33 \\
					{\it MFVI2O} & 90.53 & 92.83 & 92.12 & 81.73 & 89.72 & 92.07 & 88.53 & 92.78 & 90.19 & 91.88 & 85.88 & 92.67 & 90.07   \\ 	
					\toprule				
												\multicolumn{13}{c}{+$\text{BERT}_{\text{multilingual}}$}  \\
					\toprule
				    {\it MFVI2O} & \textbf{91.30} & 93.60 & 92.09 & 82.00 & 90.75 & 92.62 & 89.32 & 93.66 & 91.21 & 91.74 & 86.40 & 92.61  & 90.61 \\ 
					$\textit{Biaffine+MM}^{\dagger}$ & 90.30 & \textbf{94.49} & \textbf{92.65} & \textbf{85.98} & 91.13 &93.78 &\textbf{91.77} &94.72 &91.04 &94.21 &87.24 & \textbf{94.53} &91.82 \\ 
                     {\it Ours} & 91.10 & 94.46 &92.57 &85.87 & \textbf{91.32} & \textbf{93.84} & 91.69 & \textbf{94.78} & \textbf{91.65} & \textbf{94.28} & \textbf{87.48} &94.45 & \textbf{91.96} \\ 
					\bottomrule	\\		
	\end{tabular}}}}
	\caption{
		\label{ud_2.2} Labeled Attachment Score (LAS) on twelve languages in UD 2.2. We use ISO 639-1 codes to
represent languages.  $\dagger$ means our implementation. 
	}
	\label{result_ud} 
	\vskip -.12in
\end{table*}

\begin{table}[t]
    \centering 
    \small
    \begin{tabular}{lcccccccc}
        \toprule 
        & \multicolumn{2}{c}{{\bf PTB}} & \multicolumn{2}{c}{{\bf CTB}}\\
        & UAS & LAS  & UAS & LAS  \\
        \midrule 
       max-margin loss & 97.24 & 95.73 & 93.33 & 92.30 \\
       span-selection loss & 97.07 & 95.50 & 93.28 & 92.20\\
       \bottomrule 
    \end{tabular}
    \caption{The influence of training loss function on PTB and CTB.}
    \label{abl1}
\end{table}

%
%

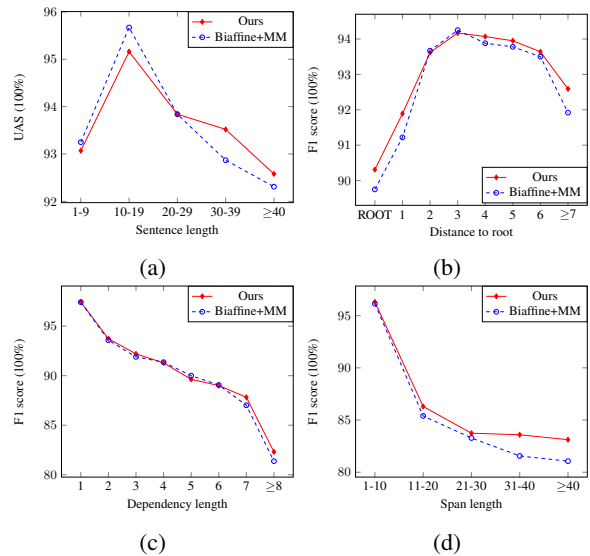
\begin{figure}[tb!]

\centering
	\begin{subfigure}[t]{0.49\linewidth}
		 \resizebox{1\textwidth}{!}{%
\begin{tikzpicture}
\begin{axis}[
    legend style={
            at={(1,1)},anchor=north east
        },
    symbolic x coords={1-9,10-19,20-29,30-39,$\ge$40},
    xtick=data,
        xlabel={Sentence length},
        ylabel={UAS (100\%)}
    ]
        \addplot[mark=diamond*,red] coordinates {
(1-9, 93.07)
 (10-19, 95.16)
 (20-29, 93.84)
 (30-39, 93.52)
 ($\ge$40, 92.58)};


    \addlegendentry{Ours}

    \addplot[mark=o,mark options={solid},blue,dashed] coordinates {
(1-9, 93.25)
 (10-19, 95.67)
 (20-29, 93.84)
 (30-39, 92.87)
 ($\ge$40, 92.31)};


    \addlegendentry{Biaffine+MM}
\end{axis}
\end{tikzpicture}
}
\caption{}
\label{error_a}
\end{subfigure}
	\begin{subfigure}[t]{0.49\linewidth}
		 \resizebox{1\textwidth}{!}{%
\begin{tikzpicture}

\begin{axis}[
    legend style={
            at={(1,.2)},anchor=north east
        },
    symbolic x coords={ROOT,1,2,3,4,5, 6, $\ge$7},
    xtick=data,
    xlabel={Distance to root},
    ylabel={F1 score (100\%)}
]

    \addplot[mark=diamond*,red] coordinates {
(ROOT, 90.31)
 (1, 91.89)
 (2, 93.62)
 (3, 94.17)
 (4, 94.07)
 (5, 93.95)
 (6, 93.64)
 ($\ge$7, 92.59)
 };

    \addlegendentry{Ours}
    \addplot[mark=o,mark options={solid},blue,dashed] coordinates {
(ROOT, 89.75)
 (1, 91.22)
 (2, 93.67)
 (3, 94.25)
 (4, 93.88)
 (5, 93.78)
 (6, 93.50)
 ($\ge$7, 91.92)
 };


    \addlegendentry{Biaffine+MM}
\end{axis}
\end{tikzpicture}
}
\caption{}
\label{error_b}
\end{subfigure}
	\begin{subfigure}[t]{0.49\linewidth}
		 \resizebox{1\textwidth}{!}{%
\begin{tikzpicture}

\begin{axis}[
    legend style={
            at={(1,1)},anchor=north east
        },
    symbolic x coords={1,2,3,4,5, 6, 7, $\ge$8},
    xtick=data,
    xlabel={Dependency length},
    ylabel={F1 score (100\%)}
]

    \addplot[mark=diamond*,red] coordinates {
 (1, 97.44)
 (2, 93.73)
 (3, 92.20)
 (4, 91.29)
 (5, 89.63)
 (6, 89.00)
 (7, 87.82)
 ($\ge$8, 82.31)
 };

    \addlegendentry{Ours}

    \addplot[mark=o,mark options={solid},blue,dashed] coordinates {
 (1, 97.42)
 (2, 93.58)
 (3, 91.90)
 (4, 91.36)
 (5, 90.00)
 (6, 89.06)
 (7, 87.01)
 ($\ge$8, 81.37)
 };


    \addlegendentry{Biaffine+MM}
\end{axis}
\end{tikzpicture}
}
\caption{}
\label{error_c}
\end{subfigure}
\begin{subfigure}[t]{0.49\linewidth}
		 \resizebox{1\textwidth}{!}{%
\begin{tikzpicture}

\begin{axis}[
    legend style={
            at={(1,1)},anchor=north east
        },
    symbolic x coords={1-10,11-20,21-30,31-40,$\ge$40},
    xtick=data,
    xlabel={Span length},
    ylabel={F1 score (100\%)}
]

    \addplot[mark=diamond*,red] coordinates {
 (1-10, 96.31)
 (11-20, 86.3)
 (21-30, 83.74)
   (31-40, 83.59)
 ($\ge$40, 83.12)
 };
    \addlegendentry{Ours}

    \addplot[mark=o,mark options={solid},blue,dashed] coordinates {
 (1-10, 96.14)
 (11-20, 85.39)
 (21-30, 83.28)
 (31-40, 81.56)
 ($\ge$40, 81.06)
 };

    \addlegendentry{Biaffine+MM}
\end{axis}
\end{tikzpicture}
}
\caption{}
\label{error_d}
\end{subfigure}

\caption{Error analysis on the CTB test set.}
\label{error_ctb}
\end{figure}

\subsection{Main result}
Table \ref{tab:ptb_ctb} shows the results on PTB and CTB. Note that Biaffine+MM is our reimplementation of the Biaffine Parser that uses the same setting as in our method, including the use of the max-margin loss instead of the local head-selection loss.  Interestingly, we find that Biaffine+MM has already surpassed many strong baselines, and this may be due to the proper choices of hyperparameters and the 
use of the max-margin loss (we observe that using the max-margin loss leads to a better performance compared with the original head-selection loss), so Biaffine+MM is a very strong baseline. It also has the same number of parameters as our methods.  Our method surpasses Biaffine+MM on both datasets, showing the competitiveness of our headed-span-based method in a fair comparison with first-order graph-based parsing. Our method also obtains the state-of-the-art result among methods that only use dependency training data (HPSG+LAL uses additional constituency trees as training data, so it is not directly comparable with the other systems.). 

Table \ref{ud_2.2} shows the results on UD. We can see that our reimplemented Biaffine+MM has already surpassed MFVI2O, which utilizes higher-order information. Our method outperforms Biaffine+MM by 0.14 LAS on average, validating the effectiveness of our proposed method in the multilingual scenarios.

\section{Analysis}

\subsection{Influence of training loss function}

Table \ref{abl1} shows the influence of the training loss function. We find that the max-margin loss performs better on both datasets: 0.17 UAS improvement on PTB and 0.05 UAS improvement on CTB comparing to the local span-selection loss, which shows the effectiveness of using global loss.

\subsection{Error analysis}
As previously argued, first-order graph-based methods are insufficient to model complex subtrees, so they may have difficulties in parsing long sentences and handling long-range dependencies. To verify this, we follow \cite{mcdonald-nivre-2011-analyzing} to plot UAS as a function of the sentence length and plot F1 scores as functions of the distance to root and dependency length on the CTB test set.  We additionally plot the F1 score of the predicted headed spans against the gold headed spans with different span lengths. 

From Figure \ref{error_a}, we can see that Biaffine+MM has a better UAS score on short sentences (of length <=20), while for long sentences (of length >=30), our headed span-based method has a higher performance, which validates our conjecture. 

Figure \ref{error_b} shows the F1 score for arcs of varying distances to root. Our model is better at predicting arcs of almost all distances to root in the dependency tree, which reveals our model's superior ability to predict complex subtrees. 
 
  Figure \ref{error_c} shows the F1 score for arcs of varying lengths. Both Biaffine+MM and our model have a very similar performance in predicting arcs of distance < 7, while our model is better at predicting arcs of distance >= 7, which validates the ability of our model at capturing long-range dependencies. 

Figure \ref{error_d} shows the F1 score for headed spans of varying lengths. We can see that when the span length is small (<=10), Biaffine+MM and our model have a very similar performance. However, our model is much better in predicting longer spans (especially for spans of length >30).

\subsection{Parsing speed}
Inspired by \citet{zhang-etal-2020-efficient} and \citet{rush-2020-torch} who independently propose to batchify the Eisner algorithm using \texttt{Pytorch}, we batchify our proposed method so that $O(n^2)$ out of $O(n^3)$ can be computed in parallel, which greatly accelerates parsing. We achieve a similar parsing speed of our method to the fast implementation of the Eisner algorithm by \citet{zhang-etal-2020-efficient}: it parses 273 sentences per second, using BERT as the encoder under a single TITAN RTX GPU. 

\section{Related work}
\paragraph{Dependency parsing with more complex subtree information.}
There has always been an interest to incorporate more complex subtree information into graph-based and transition-based methods since their invention. Before the deep learning era, it was difficult to incorporate sufficient contextual information in first-order graph-based parsers. To mitigate this, researchers develop higher-order dependency parsers to capture more contextual information \cite{mcdonald-pereira-2006-online, carreras-2007-experiments, koo-collins-2010-efficient, ma-zhao-2012-fourth}. However, incorporating more complex factors worsens inference time complexity. For example, exact inference for third-order projective dependency parsing has a $O(n^4)$ time complexity and exact inference for higher-order non-projective dependency parsing is NP-hard \cite{mcdonald-pereira-2006-online}. To decrease inference complexity, researchers use approximate parsing methods. \citet{smith-eisner-2008-dependency} use belief propagation (BP) framework for approximate inference to trade accuracy for efficiency. They show that third-order parsing can be done in $O(n^3)$ time using BP. \citet{gormley-etal-2015-approximation} unroll the BP process and use gradient descent to train their parser in an end-to-end manner. \citet{wang-tu-2020-second} extend their work by using neural scoring functions to score factors.  For higher-order non-projective parsing, researchers resort to dual decomposition algorithm (e.g., AD$^{3}$) for decoding \cite{martins-etal-2011-dual, martins-etal-2013-turning}. They observe that the approximate decoding algorithm often obtains exact solutions. \citet{fonseca-martins-2020-revisiting} combine neural scoring functions and their decoding algorithms for non-projective higher-order parsing. \citet{zheng-2017-incremental} proposes a incremental graph-based method to utilize higher-order information without hurting the advantage of global inference. \citet{ji-etal-2019-graph} use a graph attention network to incorporate higher-order information into the Biaffine Parser. \citet{zhang-etal-2020-efficient} enhance the Biaffine Parser by using a deep triaffine function to score sibling factors.
\citet{DBLP:journals/tacl/MohammadshahiH21} propose an iterative refinement network that injects the predicted soft trees from the previous iteration to the self-attention layers to
predict the soft trees of the next iteration, so that information of the whole tree is considered in parsing. As for transition-based methods, \citet{ma-etal-2018-stack, liu-etal-2019-hierarchical, DBLP:journals/corr/abs-2105-09611} incorporate sibling and grandparent information into transition-based parsing with Pointer Networks. 

\paragraph{The hook trick and the head-splitting trick.} These two tricks have been used in the parsing literature to accelerate parsing. \citet{eisner-satta-1999-efficient, eisner-satta-2000-faster} use the hook trick to decrease the parsing complexity of lexicalized PCFGs and Tree Adjoining Grammars. \citet{huang-etal-2005-machine, huang-etal-2009-binarization} adapt the hook trick to accelerate machine translation decoding. The parsing algorithms of \citet{corro-2020-span} and \citet{xin-etal-2021-n} can be viewed as adapting the hook trick to accelerate discontinuous and continuous constituency parsing, respectively.
\citet{eisner-1997-bilexical,  satta-kuhlmann-2013-efficient} use the head-splitting trick to accelerate projective and nonprojective dependency parsing. 

\paragraph{Span-based constituency parsing.} Span-based parsing is originally proposed in continuous constituency parsing \cite{stern-etal-2017-minimal, kitaev-klein-2018-constituency, DBLP:conf/ijcai/ZhangZL20, xin-etal-2021-n}. Span-based constituency parsers decompose the score of a constituency tree into the scores of its constituents.  Recovering the highest-scoring tree can be done via the exact CYK algorithm or greedy top-down approximate inference algorithm \cite{stern-etal-2017-minimal}. \citet{ kitaev-klein-2018-constituency} propose a self-attentive network to improve the parsing accuracy. They separate content and positional attentions and show the improvement. \citet{DBLP:conf/ijcai/ZhangZL20} use a two-stage bracketing-then-labeling framework and replace the max-margin loss with the TreeCRF loss \cite{finkel-etal-2008-efficient}.  \citet{xin-etal-2021-n} recently propose a recursive semi-Markov model, incorporating sibling factor scores into the score of a tree to explicitly model n-ary branching structures.  \citet{corro-2020-span} adapts span-based parsing to discontinuous constituency parsing and obtains the state-of-the-art result. 

\section{Conclusion}
In this work, we have presented a headed-span-based method for projective dependency parsing. Our proposed method can utilize more subtree information and meanwhile enjoy global training and exact inference. Experiments show the competitive performance of our method in multiple datasets. In addition to its empirical competitiveness, we believe our work provides a novel perspective of projective dependency parsing and could lay the foundation for further algorithmic advancements in the future.

\section*{Acknowledgments}
We thank the anonymous reviewers for their constructive comments. This work was supported by the National Natural Science Foundation of China (61976139).

\bibliography{anthology,custom}
\bibliographystyle{acl_natbib}

\appendix
\section{Parsing algorithm}
The parsing algorithm first computes all the chart items defined above and then recovers the parse tree from top down. For a given headed span, it finds the best segmentation of left child spans and right child spans, and then adds dependency arcs from the headword of the given headed span and the headword of each child span.  Finding the best segmentation is similar to the inference procedure of the semi-Markov CRF model \cite{DBLP:conf/nips/SarawagiC04}. Then we apply the same procedure to each child headed span (within the best segmentation) recursively.  We also maintain the following backtrack points in order to recover the predicted projective tree:
\begin{align*}
	 B_{i, j} &=     \begin{cases}
       1, \quad \alpha_{i, j} = \max\limits_{i<h\le j} (\beta_{i, j, h}) \\
       0, \quad \alpha_{i, j} = \max\limits_{i <k < j}(\alpha_{i, k} + \alpha_{k, j})
    \end{cases} \\
 \\
      C_{i, j} &= \argmax\limits_{i <k < j}(\alpha_{i, k} + \alpha_{k, j}) \\
    H_{i, j} &= \argmax\limits_{i<h\le j} (\beta_{i, j, h})
\end{align*}
The parsing algorithm is formalized in Alg.\ref{alg1}.
\begin{algorithm}[!t]
\small 
\begin{algorithmic}
\Require Input sentence of length $n$
\State Calculate all $\alpha, \beta, B, C, H$.
\State arcs $\leftarrow$ \{($\text{ROOT} \rightarrow H_{0, n})$\}
\Function{findarc}{$i, j$}
    \If{$i+1=j$}
    	\State \textbf{return} \{j\}    
    \ElsIf{$B_{i,j} = 1$}
        \State $h \leftarrow H_{i, j}$   
        \If{i+1 < h < j}
        	\State $L \leftarrow$ {\sc findarc}($i, h-1$)
        			
        	\State $R \leftarrow$ {\sc findarc}($h, j$)
        		
  	 		\State Children $\leftarrow L \cup R$ 
  	 	\ElsIf{h = j}
  	 		\State Children $\leftarrow$ {\sc findarc}($i, j-1$)
  	 	\Else
  	 		\State Children $\leftarrow$ {\sc findarc}($i+1,j$)
  	 	\EndIf
  	 	\For{c in Children}
  	 		\State $\text{arcs} \leftarrow \text{arcs} \cup (h \rightarrow c)$
  	 	\EndFor
  	 	\State \textbf{return} \{h\}
  	 \Else
  	 	\State $c \leftarrow C_{i, j}$
  	 	
  	 	\State $L \leftarrow$ {\sc findarc}($i, c$)
  	 	
  	 	\State $R \leftarrow$ {\sc findarc}($c, j$)
  	 	
  	 	\State \textbf{return} {$L \cup R$}
  	 \EndIf
  \EndFunction 
 \State {\sc findarc}($0, n$)
 \State   \textbf{return} arcs
\end{algorithmic}
\caption{Inference algorithm for headed span-based projective dependency parsing}
\label{alg1}
\end{algorithm}

\section{Implementation details}
 We use "bert-large-cased" for PTB, "bert-base-chinese" for CTB, and "bert-multilingual-cased" for UD, so the dimension of the input BERT embedding is 1024, 768, and 768 respectively. The dimension of POS tag embedding is set to 100 for CTB and UD. 
   The hidden size of BiLSTM is set to 1000. The hidden size of biaffine functions is set to 600 for scoring spans and arcs (used in our reimplemented Biaffine Parser), 300 for scoring labels. We add a dropout layer after the embedding layer, LSTM layers, and MLP layers. The dropout rate is set to 0.33. We use Adam \cite{DBLP:journals/corr/KingmaB14} as the optimizer with $\beta_1 = 0.9, \beta_2=0.9$ to train our model for 10 epochs.  The maximal learning rate is $lr=5e-5$ for BERT and $lr=25e-5$ for other components. We linearly warmup the learning rate to the maximal value for the first epoch and gradually decay it to zero for the rest of the epochs. The value of gradient clipping is set to 5. We batch sentences of similar lengths to better utilize GPUs. The token number is 4000 for each batch, i.e., the sum of lengths of sentences is 4000.

\end{document}